\documentclass[fleqn,10pt]{misc/wlscirep}
\usepackage[utf8]{inputenc}
\usepackage[T1]{fontenc}
\usepackage{colortbl}
\usepackage{nameref}
\usepackage{hyperref}
\usepackage[noabbrev,capitalize]{cleveref}
\usepackage{csquotes}
\usepackage{subcaption}
\usepackage{tabularx}
\usepackage{fancyvrb}

\usepackage[normalem]{ulem}

\title{EMMT: A simultaneous eye-tracking, 4-electrode EEG and audio corpus for multi-modal reading and translation scenarios}

\author[]{Sunit Bhattacharya}
\author[]{Věra Kloudová}
\author[]{Vilém Zouhar}
\author[]{Ondřej Bojar}

\affil[]{Institute of Formal and Applied Linguistics, Faculty of Mathematics and Physics, Charles University}

\def\dataset{Eyetracked Multi-Modal Translation (EMMT) corpus}

\begin{abstract}
We present the \dataset{}, a dataset containing monocular eye movement recordings, audio 
and 4-electrode 
electroencephalogram (EEG) data of 43 participants. The objective was to collect cognitive signals as responses of participants engaged in a number of language intensive tasks involving different text-image stimuli settings when translating from English to Czech.

Each participant was exposed to 32 text-image stimuli pairs and asked to (1) read the English sentence, (2) translate it into Czech, (3) consult the image, (4) translate again, either updating or repeating the previous translation. The text stimuli consisted of 200 unique sentences with 616 unique words coupled with 200 unique images as the visual stimuli.

The recordings were collected over a two week period and all the participants included in the study were Czech natives with strong English skills. 
Due to the nature of the tasks involved in the study and the relatively large number of participants involved, 
the corpus is well suited for research in Translation Process Studies, Cognitive Sciences 
among other disciplines.

\end{abstract}
\begin{document}

\flushbottom 
\maketitle
\thispagestyle{empty}

\section{Background \& Summary}
Rapid advances in Artificial Intelligence over the last few years have translated to significant advances in Natural Language Processing (NLP).
The extraordinary effectiveness of the neural network approach for artificial intelligence is seen in the performance of the state-of-the-art language models\cite{dale2021gpt} and in the ways that they are being fine-tuned for different NLP applications. The state-of-the-art machine translation systems have been reported to have achieved news translation quality comparable to human professionals.\cite{popel2020transforming} Inspired by these successes, machine translation research in the last few years has also attempted to target the problem of multimodal translation.\cite{Sulubacak2020multimodalmt}

While traditional translation is thought of as being about printed words,\cite{o2013introduction} the rise of the Internet and new media have made multimodal texts ubiquitous. This implies that there is a need for a systematic understanding of translation processes for multimodal texts. This understanding, we posit, should arise from learning how both human and machine ``brains'' function. The discipline of artificial intelligence has been historically `conceptualized in anthropomorphic terms'. \cite{watson2019rhetoric} And although there are many risks and incorrect assumptions when anthropomorphizing modern deep neural networks,\cite{funke2021five} there has been a growing body of work analyzing this very aspect of such systems.
One reasoning behind such an analysis is that a comparative analysis of human and machine intelligence should lead to a better understanding of intelligence. There is a view that language is a key to understanding human intelligence.\cite{premack2004language} It is thus fair to assume that cognitive signal data from a participant performing a set of tasks involving language use would be a great resource to understand how the brain works with language.
If the tasks are designed carefully such that they can also be presented to trained deep neural network algorithms,
they can be used to compare 
the human and machine perspective of handling the same problem, see e.g. the recent comparison of interpreting occluded scenes.\cite{murlidaran2021comparing} A similar approach in computer vision has led to simpler and yet equally effective models for object detection.\cite{brainlike2019}

Our objective is to collect cognitive data in the form of gaze and EEG signals from participants while they translate sentences in multimodal conditions. We hope that this dataset would help researchers study how behavioural patterns (in terms of cognitive signals) change with stimuli of differing modalities.
The tasks that the participants are subjected to have been designed so that similar experiments can be performed on the state-of-the-art algorithms for natural language processing.\cite{Sulubacak2020multimodalmt}

Whether the observed eye-tracking and EEG data are sufficient to relate human and machine processing of text accompanied by an image is an empirical question.
Over the years, many eye-tracking studies have been successfully done for translation process research, leading to key insights.\cite{doherty2010eye,stymne2012eye,moorkens2018eye,doherty2014assessing}
In recent years, many studies involving EEG signals, too, have been conducted to understand the neural underpinnings of language.\cite{ganushchak2011use}
Our dataset will hopefully foster further research in that direction. 

\section{Methods}
\subsection{Equipment Used}
\begin{figure}[hb]
\centering
\includegraphics[width=0.4\textwidth]{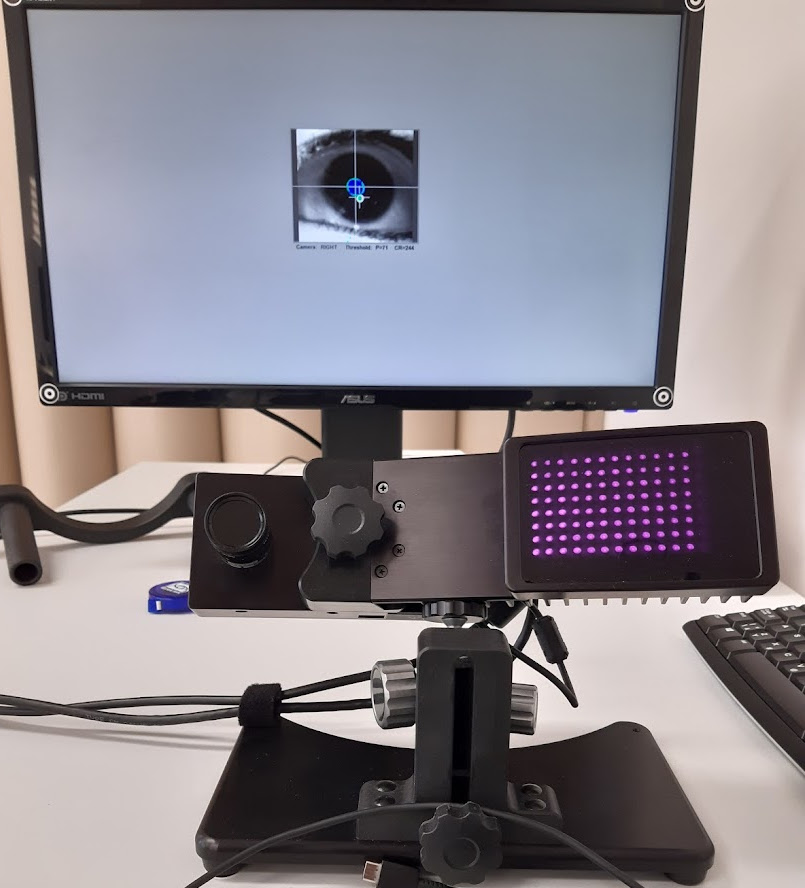}
\caption{Experiment setup with eyetracker (EEG and audio recorder not shown).}
\label{fig:exp_setup}
\end{figure}

An EyeLink 1000 Plus by SR-Research was used for recording the gaze data. The eye-tracker was kept at a distance of 45 cm from the participants. The screen where the stimuli were presented was kept at 65 cm from the participants. The experimental setup from the perspective of the participants is shown in \Cref{fig:exp_setup}. 
The Muse 2\footnote{\href{https://choosemuse.com/muse-2/}{choosemuse.com/muse-2}} headband was used to record the EEG data from the experiments.
Low-cost portable headbands like Muse have been recently used for a number of ERP-based (Event-Related Potentials\cite{luck2012event,woodman2010brief}) studies with reasonable results.\cite{krigolson2017choosing} Along with this, the audio of the entire session was recorded using a Samsung Galaxy M21 device.

\subsection{Participants}
Data were recorded from 44 people who volunteered to be a part of the study. Requests for participants to take part was circulated via flyers, emails, tweets and word of mouth.
Out of the 44 people who volunteered, data from 43 people were used in the dataset.
All of the participants were native speakers of Czech (in the age group between 17 and 51) who also spoke English as a second or third language.
The distribution of the age of participants is shown in \Cref{table:age_part}.

The participants were asked to fill out a self-reporting questionnaire to gather additional data. That data included facts like the level of proficiency in English and if the participant offered translation services professionally.  Information about the distribution of the level of proficiency in English for the participants (on the CEFR\footnote{\href{https://www.coe.int/en/web/common-european-framework-reference-languages/level-descriptions}{coe.int/en/web/common-european-framework-reference-languages/level-descriptions}} scale) is given in \Cref{table:eng_proficiency}. 

\begin{table}[ht]
\centering
\begin{tabular}{lcccccccc}
\hline
    \toprule
    Age & 17-20 & 21-25 & 26-30 & 31-35 & 36-40 & 41-45 & 45-50 & 51-55\\
    \midrule
    Participants & 5 & 21 & 8 & 2 & 4 & 1 & 1 & 1 \\
    \bottomrule
\end{tabular}
\caption{Distribution of the participants' age.}
\label{table:age_part}
\end{table}

\begin{table}[ht]
\centering
\begin{tabular}{lcccc}
\hline
    \toprule
    CEFR Level & B1 & B2 & C1 & C2 \\
    \midrule
    Participants & 2 & 15 & 16 & 8 \\
    \bottomrule
\end{tabular}
\caption{Level of proficiency in English of the participants on the CEFR scale.}
\label{table:eng_proficiency}
\end{table}

\begin{table}[ht]
\centering
\begin{minipage}{0.39\textwidth}
    \begin{tabularx}{\textwidth}{lcc} 
    \toprule
    & Professional & Amateur \\ 
    \midrule
    Participants & 16 & 27 \\
    \bottomrule
    \end{tabularx}
    \caption{Distribution of participants providing language services (translation or interpretation) professionally.}
    \label{table:amatueur_prof}
\end{minipage} %
\hspace{0.5cm}
\begin{minipage}{0.5\textwidth}
    \begin{tabularx}{\textwidth}{lccc}
    \toprule
    Service & Occasionally & 1-5 years & >10 years \\ 
    \midrule
    Translation & 11 & 0 & 1 \\
    Interpretation & 10 & 3 & 1\\
    \bottomrule
    \end{tabularx}
    \caption{Self-reported experience of participants in providing either translation or interpretation services.}
    \label{table:prof_trans_inter}
\end{minipage}
\end{table}

\Cref{table:amatueur_prof} shows the number of people self-reporting to be amateurs and professionals (those who provide translation and interpretation services professionally). \Cref{table:prof_trans_inter} shows the distribution of people self-reporting their experience in providing translation and interpretation services respectively. The cognitive signals from participants with such diversity of professional experience in providing language services could provide interesting insights into how language processing strategies vary across people.

The experiments were supervised across a span of 12 by two proctors (authors of this paper).
The experiments were designed keeping in mind the code of ethics of Charles University.\footnote{Available in Czech: \href{https://cuni.cz/UK-9490.html}{cuni.cz/UK-9490.html}. The experiment is further GDPR-compliant and approved by a qualified university staff \href{mailto:kapralova@ufal.mff.cuni.cz}{Libuše Kaprálová}.}
All the participants were duly informed about the tasks involved in the experiment before the beginning of the experiment.
The participants in the study also signed a consent form agreeing to the recording of the gaze, audio EEG data together with the matching questionnaire answers and releasing it for research purposes.
The collected data were carefully stripped of all personal information.\footnote{Internal records were pseudo-anonymized, actual data do not contain any identifiable personal information.}

\subsection{Materials} 

The dataset that we present is composed of sentences drawn from three well-known datasets used in various machine learning tasks: Ambiguous COCO,\cite{elliott2017findings} LAVA\cite{berzak2016you} and Hindi Visual Genome (HVG).\cite{parida2019hindi}
All three corpora are multimodal corpora with an image associated with every sentence. Furthermore, all three corpora are publicly available. The original datasets have all been released with licenses that allow reuse of the images for research purposes. 

The sentence selection process initially involved annotators identifying ambiguous sentences. Accordingly, 176 sentences were marked as being `ambiguous'.\footnote{Here we do not explicitly differentiate between ambiguity and vagueness. Many `ambiguous' sentences in the dataset might be considered vague.} Two different annotators then further ranked the 176 sentences for the `level of ambiguity' in the sentence on a scale of 1 to 5. The top 100 sentences from the ranked list were selected for the experiment. Another 100 sentences from the three corpora (i.e. avoiding the 176 initially chosen by annotators) were then chosen to serve as contrastive, `unambiguous', sentences.
It was however found that the chosen sentences from the original datasets had many grammatical errors. Thus, another annotator (a native English speaker) proofread and corrected all the selected sentences.

The sentences are complemented with images as described in the Stimuli section below.
All images always come from the same corpus as the respective sentence.

\subsection{Stimuli \& Experiment Design}
For each participant, the experiment consisted of the presentation of 30 different stimuli, each based on an image-text probe.
Each probe had four sequential stages which we denote as \texttt{READ}, \texttt{TRANSLATE}, \texttt{SEE} and \texttt{UPDATE}.
The description of the stages is provided later in the section.
While the text was presented to the participants in all the stages, the image in each stimulus was presented only for the \texttt{SEE} and \texttt{UPDATE} stages.

In total, we prepared 600 different stimuli across 2$\times$3 conditions and divided them randomly into 20 probes (constrained by not showing the same sentence more than once), each containing 30 stimuli.
With 43 participants, the objective was to have at least two sets of observations for each probe.

The 2$\times$3 conditions were: 100 ambiguous (A) and 100 unambiguous (U) sentences, as described above, each complemented with a congruent (C), incongruent (I) and missing (M) image.
Congruent images are those where the image corresponds to the original sentence from the given source corpus.
Incongruent images were selected randomly from the same corpus as the sentence, avoiding the already selected congruent images. The image thus matches the style of the corpus but it is unrelated to the sentence. Finally, 100+100 stimuli contained the ``missing'' image, i.e. an image with the fixed text \textit{No visual clue for this case!}.

%

\begin{figure}[htbp]
    \centering
    \includegraphics[scale=0.8]{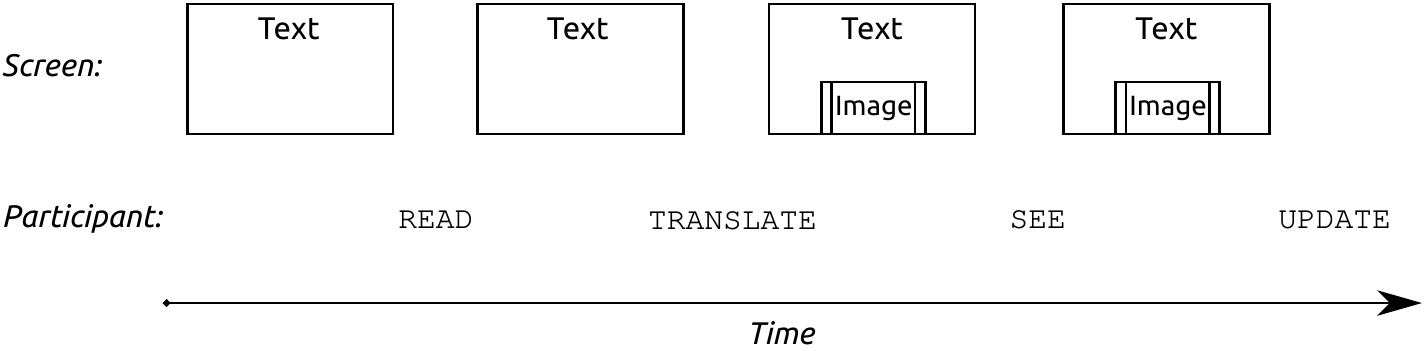}
    \caption{Visualization of the four experiment stages.}
    \label{fig:time_arch}
\end{figure}

The text was presented with a black font of font size 28 (Free Sans Bold) on a light grey background. This resulted in a letter size of approximately 9.9\,mm, i.e. 0.015$^{\circ}$ (the distance between the screen and the participants was 65\,cm). The longest sentence presented contained 97 letters (21 words) while the shortest contained 19 letters (4 words). All the sentences were presented in a single line. The participants used the spacebar key to proceed with the experiment.
All the stages were self-paced.
Each press of the spacebar was confirmed with a short audio signal (a beep), thus indicating the progress in the experiment. There were 4 example stimuli before the start of the experiment to get the participants acquainted with the format of the experiment.
The description of the stages involved in the experiment is given below and schemed in \Cref{fig:time_arch}.

\begin{figure}[ht]
\centering
     \centering
     \begin{subfigure}[b]{0.49\textwidth}
         \centering
         \includegraphics[width=\textwidth]{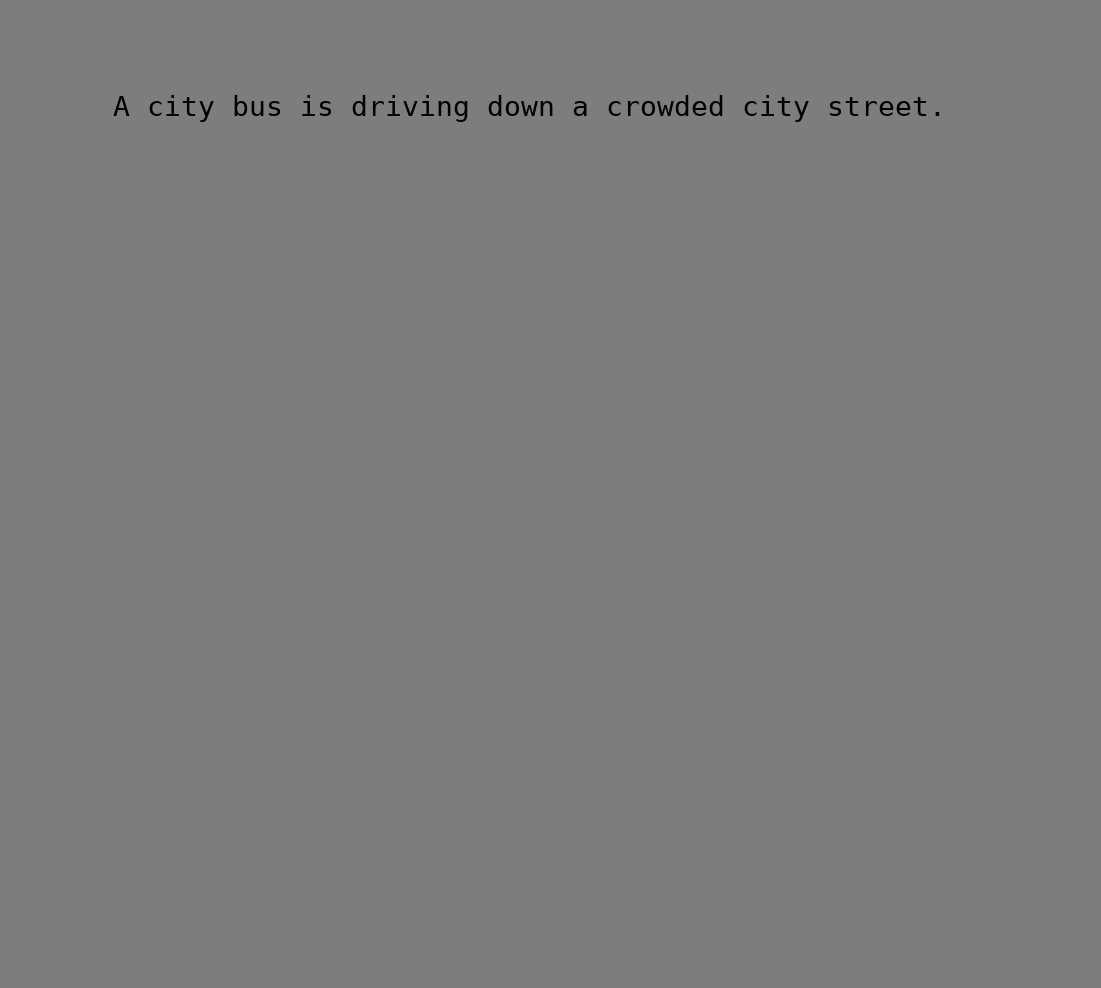}
         \caption{\texttt{READ} and \texttt{TRANSLATE} stages.}
         \label{fig:read_screen}
     \end{subfigure}
     \hfill
     \begin{subfigure}[b]{0.49\textwidth}
         \centering
         \includegraphics[width=\textwidth]{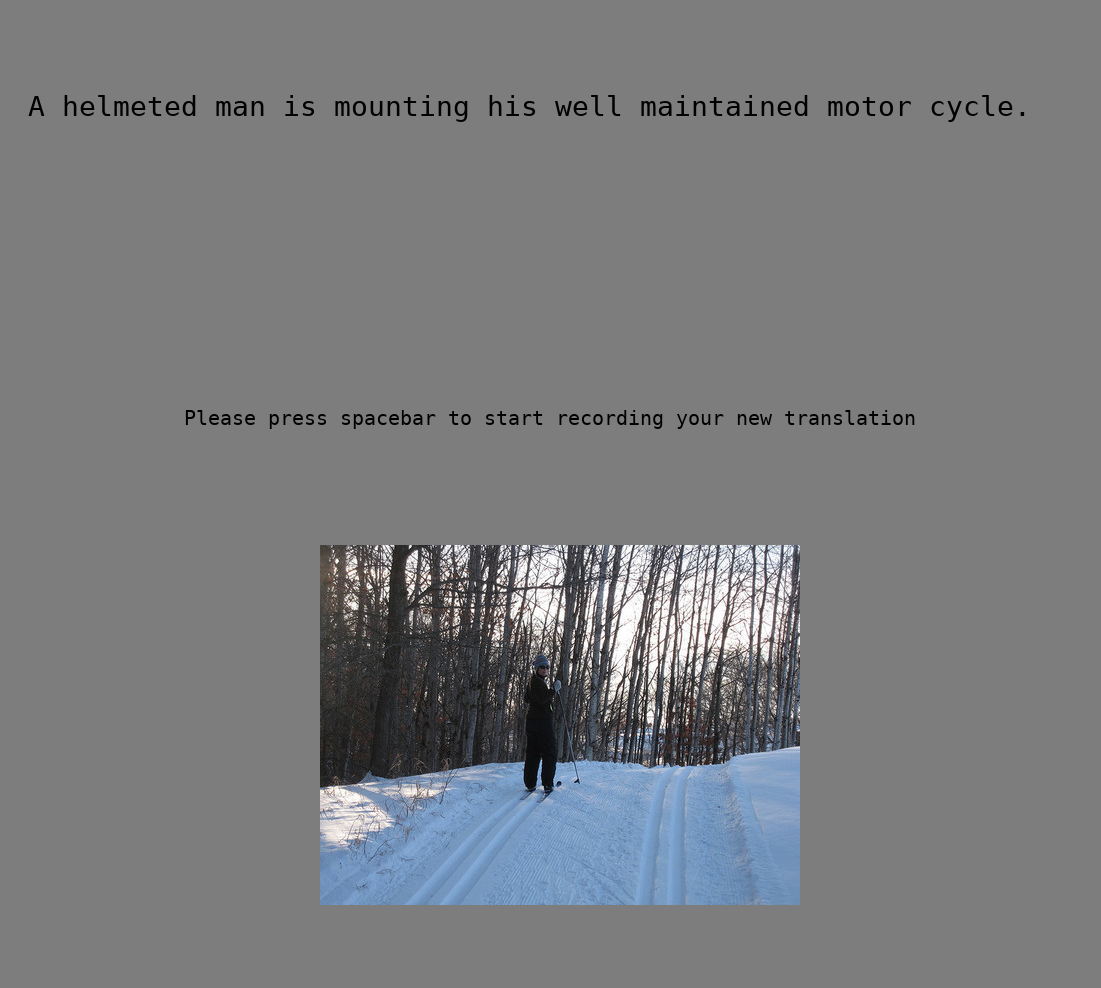}
         \caption{\texttt{SEE} and \texttt{UPDATE} stages.}
         \label{fig:see_screen}
     \end{subfigure}
\caption{Example screens (cropped) for two different stimuli.}
\label{fig:example_screens}
\end{figure}

\subsubsection{Stage 1: Read}
In the \texttt{READ} stage, the participants were shown a sentence on the display screen.
The stage involved loud reading of the sentence (in English).
An example of the screen during this stage is shown in \Cref{fig:read_screen}.
The participants were instructed to press the space bar once they were ready to move on to the next stage. 

\subsubsection{Stage 2: Translate}
The \texttt{TRANSLATE} stage used the same screen as \texttt{READ}.
The participants were expected to translate the English sentence into Czech, speaking it out loud.
Upon completion of the translation, the participants pressed the space bar, indicating the end of the stage.

\subsubsection{Stage 3: See}
Upon entering the \texttt{SEE} stage, the image (or the ``missing'' image) was presented below the sentence.
The participant was expected to look at the image and decide if they would like to change the translation they produced in the previous stage.
When decided, the participant verbally indicated that they would either like to change the translation or keep the same translation.
Pressing the space bar then led the participant to the final stage.
An example screen for this stage is shown in \Cref{fig:see_screen}.

\subsubsection{Stage 4: Update}
This stage involved the participants either repeating the translation that they made during \texttt{TRANSLATE} or updating their translation based on the visual stimuli during \texttt{SEE}.
The idea is that the participants may decide to integrate more information or correct an incorrectly guessed ambiguity based on the image (especially for congruent images) in their translation.
The collected cognitive data would hopefully reflect how that integration manifests during \texttt{SEE} and \texttt{UPDATE}.
Inevitably, the participants will not have remembered their original translation exactly, so we expect many ``unchanged'' translations to actually differ.

\subsection{Procedure}
Data were recorded from each participant in a single session.
Each experiment started with the calibration and validation of the equipment involved (eye-tracker and EEG recorder).
Each participant was then led through a practice round with four dummy stimuli, to get them acquainted with the procedures of the experiment.
Being a self-paced experiment design, the participants were given an option to temporarily pause the experiment after completing the four stages of a stimulus to take a small break.
If the participants opted to pause, the experiment would resume again with calibration and validation before starting from where it was stopped.

\section{Data Records}
The \dataset{} is hosted on the Center for Open Science (OSF) repository: {\href{https://osf.io/hxymj/}{osf.io/hxymj}}.
It is distributed under the Creative Commons License CC0 version 4.0.
All data has been anonymized in accordance with the consent provided by participants and abiding by the GDPR policies.
We segmented the recordings based on stimuli, leading to a total of 1169 audio files. The dataset also includes EEG recordings from 28 participants and log files and gaze data corresponding to 43 participants.
No data were excluded but the EEG failed for some participants, resulting in only 702 fully valid EEG recordings.

\section{Technical Validation}
In this section, we provide a small-scale visualization to better understand the data structure and its quality.
We also list several experiment applications for this data.

\subsection{Audio}

Each session audio is recorded together with stage timestamps. 
The audio analysis can serve as a probe to translation difficulty or surprisal, e.g pauses in hard-to-translate parts or prosody.
It can also help in understanding human text to speech translation characteristics.

\subsection{Eyetracking}

The data were sampled at 20kHz.\footnote{Line format: \texttt{timestamp, x-location, y-location, pupil dilation}.}
Following the annotator gaze is important in determining the attention level and location.
This is crucial not only for determining which parts of the image were important for the translation but also helps us in determining the sequence in which the screen was processed (e.g. \textit{Was the sentence read linearly or with saccades?} \textit{Did the annotators prefer to first read the whole sentence again and then look at the image or did they jump between the two areas?}).

\begin{figure}[ht]
\centering
     \centering
     \begin{subfigure}[b]{0.49\textwidth}
         \centering
         \includegraphics[width=\textwidth]{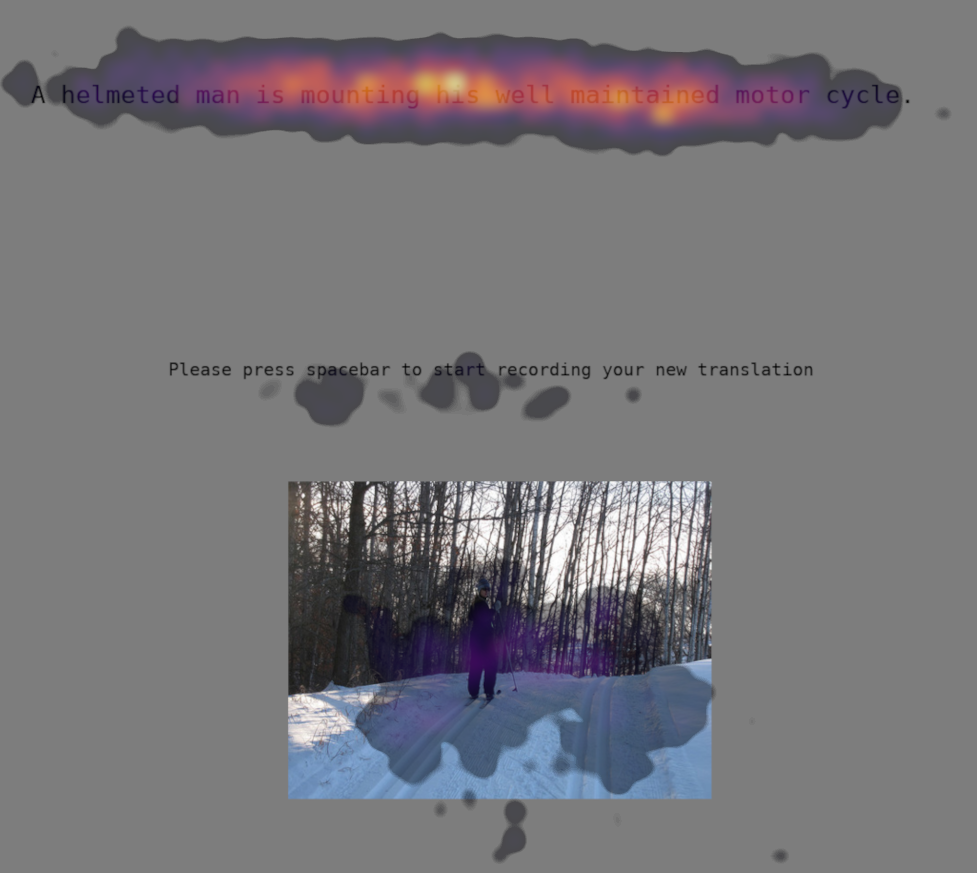}
         \vspace{0.16cm}
         \caption{Heatmap of gaze frequency (averaged over multiple annotators).\newline}
         \label{fig:et_heatmap}
     \end{subfigure}
     \hfill
     \begin{subfigure}[b]{0.49\textwidth}
         \centering
         \includegraphics[width=\textwidth]{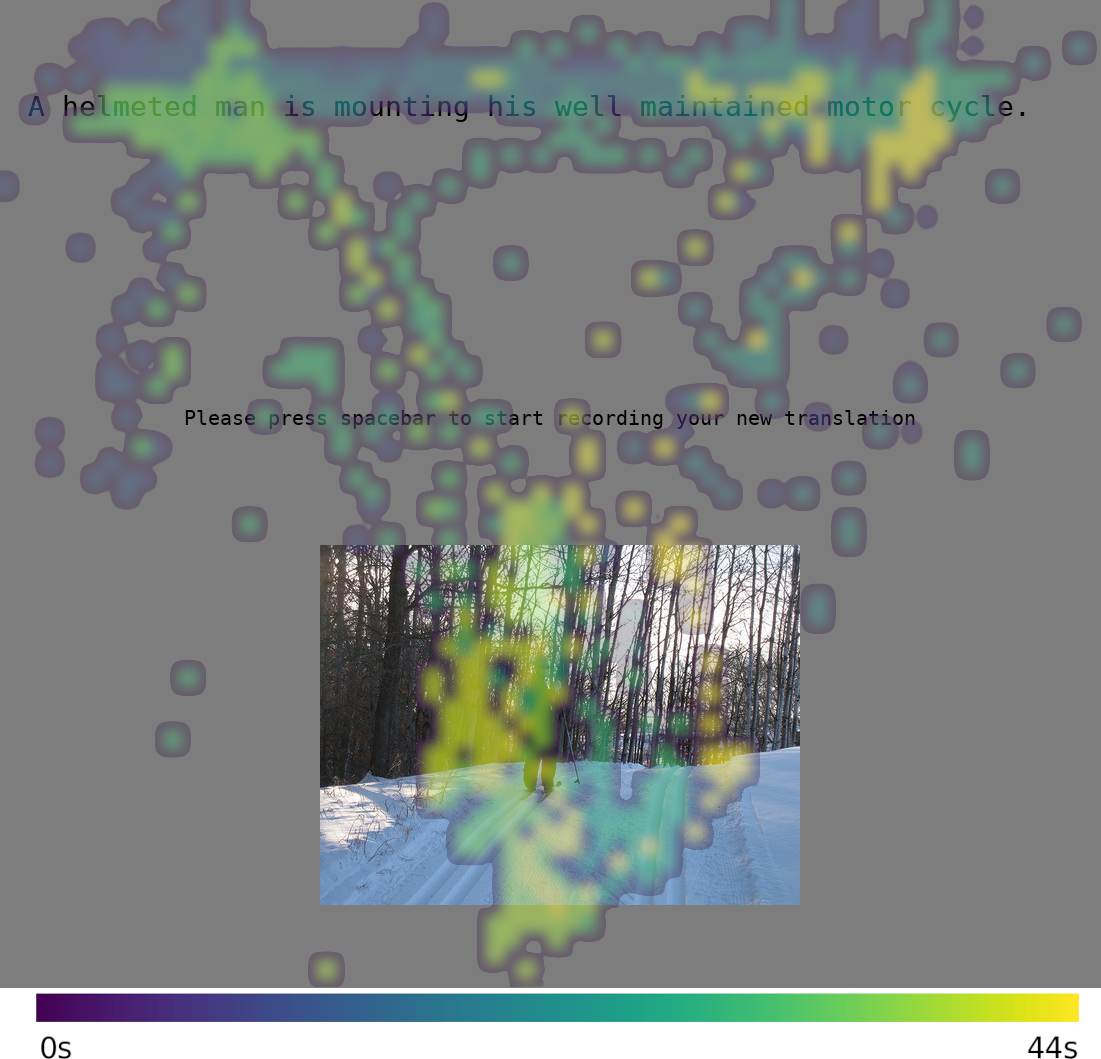}
         \caption{Gaze progression of a single participant for a specific probe. Old gaze points are occluded by newer.}
         \label{fig:et_sequence}
     \end{subfigure}
\caption{Eyetracking visualization.}
\label{fig:et_examples}
\end{figure}

\Cref{fig:et_heatmap} shows the heatmap of annotators' gaze (averaged over multiple annotators) which reveals which parts of the sentence were important/on what parts the annotators focused.
\Cref{fig:et_sequence} shows the whole gaze sequence for a single participant with colour-coded time progression.
It shows, for example, which parts of the sentence had to be consulted with the image.

\subsection{EEG}

\begin{figure}[ht]
\centering
     \centering
     \begin{subfigure}[b]{0.49\textwidth}
         \centering
         \includegraphics[width=\textwidth]{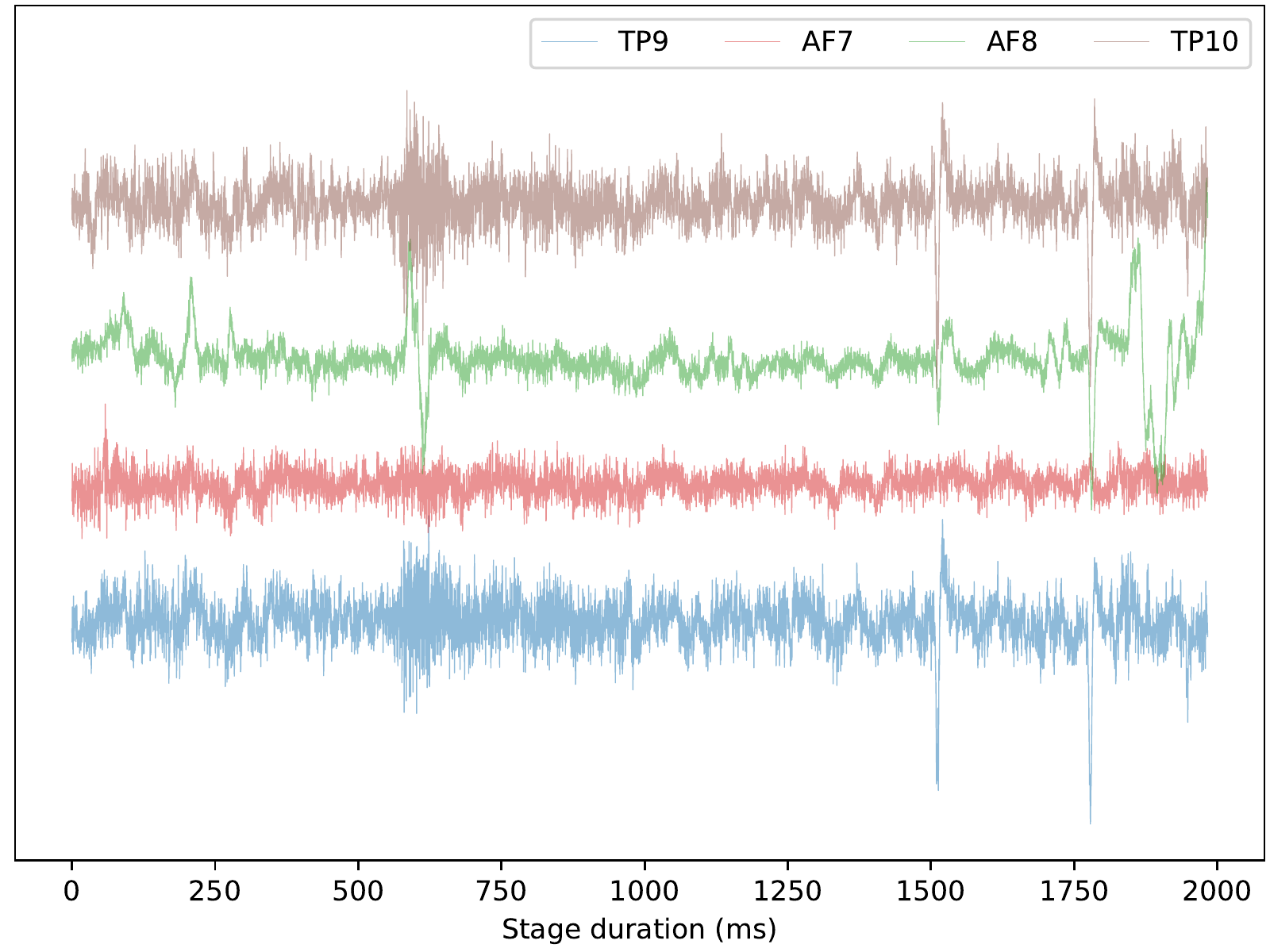}
         \caption{Raw EEG signal for all four nodes, measured in uV, of the whole first stage. They are separated for visibility (otherwise overlapping) and the y-axis is hidden.}
         \label{fig:eeg_raw}
     \end{subfigure}
     \hfill
     \begin{subfigure}[b]{0.49\textwidth}
         \centering
         \includegraphics[width=\textwidth]{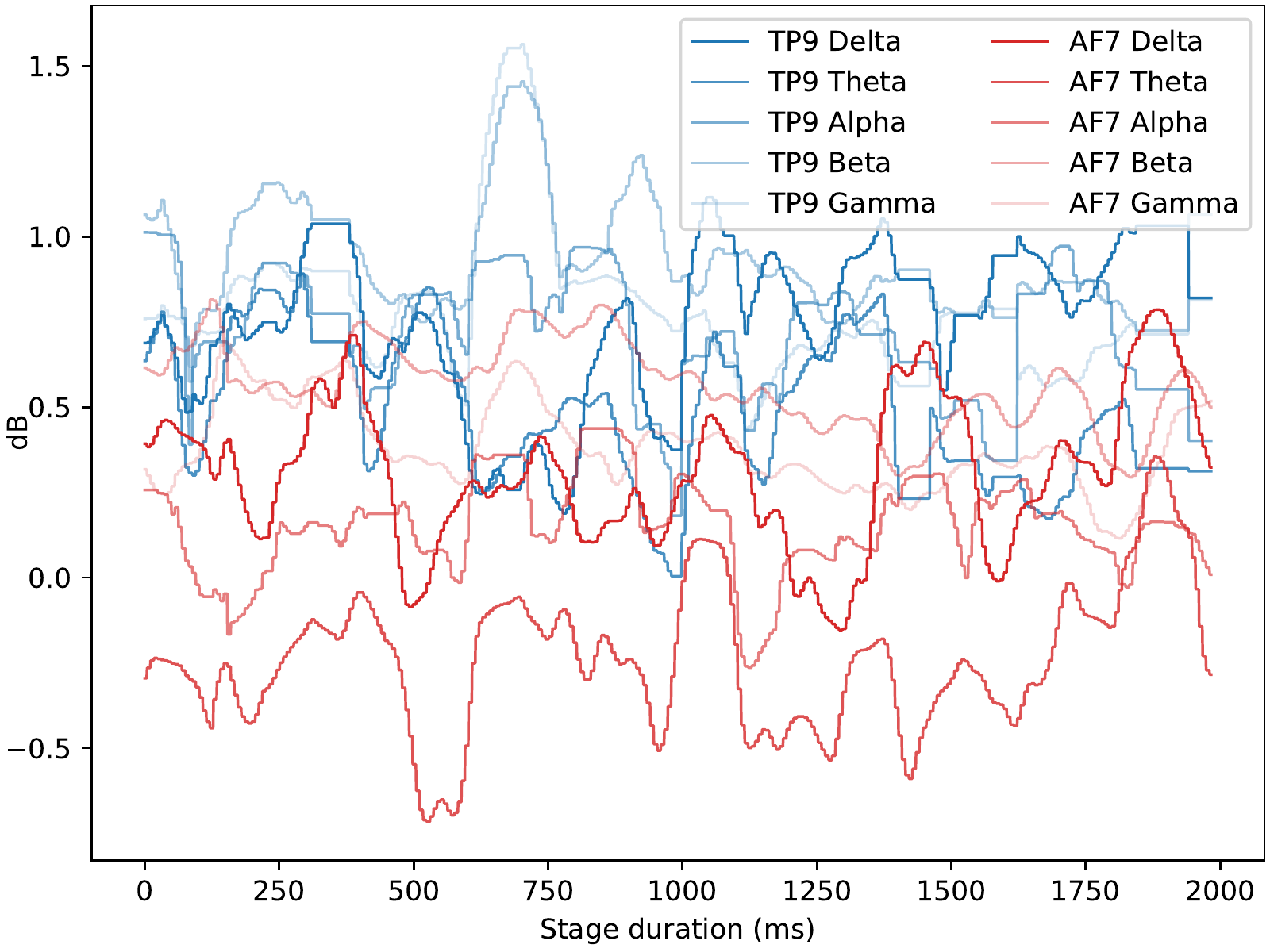}
         \caption{Four precomputed power bands for two nodes (others hidden) of the whole first stage.\newline}
         \label{fig:eeg_bands}
     \end{subfigure}
\caption{EEG recordings of the first stage of a single participant.}
\label{fig:eeg_examples}
\end{figure}

The data contains the recorded EEG signal of 4 nodes: TP9, AF7, AF8 and TP10 (see precise description of the placement\cite{bird2019deep}) sampled at 256Hz.
A more convenient view of the data is its dominant frequencies (using the fast Fourier transform).
The bands are established as: Delta: 1-4Hz, Theta: 4-8Hz, Alpha: 7.5-13Hz, Beta: 13-30Hz, Gamma: 30-44Hz.
The headband also records accelerometer data and certain muscle movements (blinks and jaw clenches).\footnote{Line format: \href{https://web.archive.org/web/20181105231756/http://developer.choosemuse.com/tools/available-data}{web.archive.org/web/20181105231756/http://developer.choosemuse.com/tools/available-data}}
The two figures in \Cref{fig:eeg_examples} show the EEG of the first stage of a specific sentence \& participant combination either as the power spectrum or as the raw EEG voltages.


Many psycholinguistic experiments make use of Event-Related Potentials: a brain response in a specific place after exposure to a stimuli (after a specific time period).\cite{aurnhammer2019evaluating,frank2015erp,brouwer2021neurobehavioral,tabullo2012eeg,bhattasali2020alice}
This EEG data allows researchers to explore the brain response to ambiguity in the task of translation with and without multimodal (image) cues.
Future experiments may include analysing the signals and seeing if it contains enough information to determine whether the sentence was ambiguous or whether the image was related.

\section{Usage Notes}

The provided data is stored in the following structure (two directories).
The \texttt{probes/Sentences.csv} file maps sentences ids to the sentence texts. The \texttt{probes/Participants.csv} file maps participant ids to the pseudo-anonymous names used by the participants for the experiment.
The \texttt{*.eeg}, \texttt{*.et} and \texttt{*.mp3} files contain the recorded EEG, eyetracking and audio data, respectively.
The \texttt{*.events} file contains timestamps of when the text was presented, the recording was started, the image was presented etc.
They all belong to a single sentence + ambiguity configuration + annotator tuple.
The sentence and ambiguity for all the runs in the \texttt{probe*} folder is listed in \texttt{probe*/probes}.

\medskip

\begin{center}
\begin{minipage}{0.4\linewidth}
\begin{Verbatim}[fontsize=\small]
probes/
  ...
  probe05/
    ...
    P10-00-S182-A-C-ia018.eeg
    P10-00-S182-A-C-ia018.et
    P10-00-S182-A-C-ia018.events
    P10-00-S182-A-C-ia018.mp3
    ...
    probes
  ...
  Sentences.csv
  Participants.csv
\end{Verbatim}
\end{minipage} 
\begin{minipage}{0.4\linewidth}
\begin{Verbatim}[fontsize=\small]
images/
  ...
  a006.jpg
  a007.jpg
  ...
  
  
  
  
  
  
  
  
\end{Verbatim}
\end{minipage}
\end{center}

\medskip

This storage format was chosen to be easily globbed (e.g. \texttt{probe*/*-S112-*.eeg} yields the sentence 112 across all annotators).
Working directly with so many files will be slow on most filesystems and we, therefore, recommend preprocessing and saving only the relevant view of the data (e.g. only the EEG of the first stage of ambiguous sentences) and saving it in a binary format, such as pickle,\footnote{\href{https://docs.python.org/3/library/pickle.html}{docs.python.org/3/library/pickle.html}} which is easier and faster to load.\footnote{We avoided using this format for distribution for forward compatibility and to not publish a large binary blob.}

\section{Code availability}
The source code for the recording of the experiment using PyGaze is available open-source online {\href{https://github.com/NEUREM3/recording-code-for-eyetracked-multi-modal-translation}{github.com/NEUREM3/recording-code-for-eyetracked-multi-modal-translation}}.
The data itself is also accessible on GitHub {\href{https://github.com/ufal/eyetracked-multi-modal-translation}{github.com/ufal/eyetracked-multi-modal-translation}}.

\bibliography{misc/bibliography.bib}

\section{Acknowledgements} 
This work has been funded from the grants  19-26934X (NEUREM3) of the Czech Science Foundation and H2020-ICT-2018-2-825303 (Bergamot) of the European Union. 
The work has also been supported by the Ministry of Education, Youth and Sports of the Czech Republic, Project No. LM2018101 LINDAT/CLARIAH-CZ.

\section{Author contributions statement}
S.B. and O.B conceived the experiment and all authors contributed to the experiment design.
S.B, V.Z and O.B designed and wrote the code.
S.B. and V.K. conducted the experiment.
All authors wrote and reviewed the manuscript. 

\section{Competing interests}
I declare that the authors have no competing interests as defined by Nature Research, or other interests that might be perceived to influence the results and/or discussion reported in this paper.

\end{document}